\documentclass[10pt,twocolumn,letterpaper]{article}

\usepackage{cvpr}
\usepackage{times}
\usepackage{epsfig}
\usepackage{graphicx}
\usepackage{amsmath}
\usepackage{amssymb}
\usepackage{lipsum}
\usepackage{comment}
\usepackage{todonotes}
\usepackage{tikz}
\usetikzlibrary{shapes.geometric, arrows}
\usetikzlibrary{decorations.pathreplacing}
\tikzstyle{arrow} = [thick,->,>=stealth]

\usepackage[pagebackref=true,breaklinks=true,letterpaper=true,colorlinks,bookmarks=false]{hyperref}
\usepackage[nameinlink, noabbrev]{cleveref}

\cvprfinalcopy 

\def\cvprPaperID{12} 

\ifcvprfinal\fi

\newcommand{\commentMR}[1]{\textcolor{black}{#1}}
\newcommand{\commentHG}[1]{\textcolor{black}{{#1}}}

\newcommand{\commentRC}[1]{\textcolor{black}{{#1}}}
\newcommand{\commentRD}[1]{\textcolor{black}{{#1}}}

\newcommand{\argmax}[1]{\underset{#1}{\operatorname{argmax}}\;}
\newcommand{\argmin}[1]{\underset{#1}{\operatorname{argmin}}\;}

\begin{document}

\title{The Ethical Dilemma when (not) Setting up Cost-based Decision Rules\\ in Semantic Segmentation}


\author{Robin Chan$^{1}$, Matthias Rottmann$^{1}$, Radin Dardashti$^{2}$,\\ Fabian H\"uger$^{3}$, Peter Schlicht$^{3}$ and Hanno Gottschalk$^{1}$ \\
\\
$^{1}$University of Wuppertal, School of Mathematics and Natural Sciences\\
$^{2}$University of Wuppertal, Philosophical Seminar,  Philosophy of Science\\
$^{3}$Volkswagen Group Research, Automated Driving, Architecture and AI Technologies\\
\tt\small{\{\href{mailto:rchan@uni-wuppertal.de}{rchan},\href{mailto:rottmann@uni-wuppertal.de}{rottmann},\href{mailto:dardashti@uni-wuppertal.de}{dardashti},\href{mailto:hgottsch@uni-wuppertal.de}{hgottsch}\}@uni-wuppertal.de}\\
\tt\small{\{\href{mailto:peter.schlicht@volkswagen.de}{peter.schlicht},\href{mailto:fabian.hueger@volkswagen.de}{fabian.hueger}\}@volkswagen.de} 
}


\maketitle
\thispagestyle{empty}

\begin{abstract}
Neural networks for semantic segmentation can be seen as statistical models that provide for each pixel of one image a probability distribution on predefined classes. The predicted class is then usually obtained by the maximum a-posteriori probability (MAP) which is known as Bayes rule in decision theory. From decision theory we also know that the Bayes rule is optimal regarding the simple symmetric cost function. Therefore, it weights each type of confusion between two different classes equally, e.g.,~given images of urban street scenes there is no distinction in the cost function if the network confuses a person with a street or a building with a tree.
Intuitively, there might be confusions of classes that are more important to avoid than others. In this work, we want to raise awareness of the possibility of explicitly defining confusion costs and the associated ethical difficulties if it comes down to providing numbers. \commentRC{We define two cost functions from different extreme perspectives, an egoistic and an altruistic one,} and show how safety relevant quantities like \commentRC{precision / recall and (segment-wise) false positive / negative rate} change when interpolating between MAP, egoistic and altruistic decision rules.
\end{abstract}

\section{Introduction} \label{sec:intro}


Machines acting autonomously in spaces co-populated by humans and robots are no longer a futuristic vision, but are part of the agenda of the world's technologically most advanced corporations. Autonomous car driving has seen spectacular advances due to recent progress in artificial intelligence (AI) and therefore is one of the corner-cases for this development. As street traffic, according to the world health organization (WHO), causes an annual death toll of 1.35M persons at the time of writing  \cite{WHO}, it \commentRD{is expected} that also autonomous driving cars will be involved in such tragic events. \commentRD{While there are reasons to believe that autonomous driving can reduce the overall numbers of deaths and heavy injuries, besides being required by \emph{e.g.}~the Ethics Commission instated by the German Federal Ministry of Transport and Digital Infrastructure \cite{Ethikkommission},  many further ethical issues remain in the choices of programming an autonomous vehicle. Therefore, autonomous cars have been a much-discussed topic in robot ethics \cite{lin2017robot}, ranging from inevitable ethical dilemmas like the trolley problem \cite{foot5problem,lin2016ethics} to more mundane ethical situations \cite{himmelreich2018never}.} 

\begin{figure*}
\centering
\input{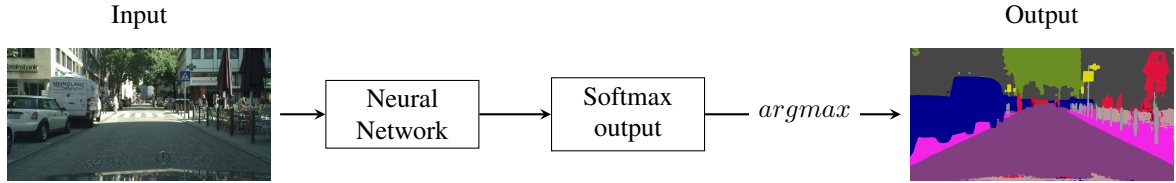}
\caption{Illustration of semantic segmentation performed on an image of the Cityscapes dataset \cite{cityscapes16} with a neural network in combination with (pixel-wise) maximum a-posteriori probability classification.}
\label{fig:sem-seg}
\end{figure*}

\commentRD{In most of these ethical situations discussed in the literature, the} robots and the AI algorithms controlling them are assumed to know the situation they decide on, whereas most deadly accidents with the involvement of self-driving cars in some way or another are connected with the (insufficient) perception of the vehicle's surrounding (see \cite{accident} for a preliminary report). \commentRD{Whether} the AI algorithms of perception themselves depend on choices that involve ethical decisions is therefore a legitimate question.

For a practitioner in the field it is quite obvious that the answer is ``yes'': In semantic segmentation, the choice of training data, the selection of classes, potential class imbalance, the amount of data, the capacity of the learning algorithm and the performance of the hardware all determine what a contemporary AI algorithm is able to ``see'' and how error prone its perception will be. As errors in perception are potential root causes of accidents, ethical implications clearly exist. 

In this work, we draw the attention to one further issue that is connected to the probabilistic output of semantic segmentation neural networks that are mostly used for the perceptive task. As the softmax output of a segmentation network gives a pixel-wise class distribution, the maximum a-posteriori probability (MAP) principle, also known as Bayes decision rule, selects the class of highest probability. This is however not the only selection principle, as one could also apply the Maximum Likelihood (ML) decision rule that picks the class for which the input data is most representative \cite{Fahrmeir-1996}. While both rules have the appeal of being mathematically ``natural'', they are merely two examples of cost-based decision rules, where each confusion event is penalized by a specific quantity $c(\hat k,k)$ that valuates the aversion of a decision maker towards the confusion of the predicted class $\hat k$ with the actual class $k$. The decision on the predicted class now minimizes the expected cost. 

Seen from this angle, the MAP principle corresponds to the cost matrix that attributes equal cost to any confusion event. We call this the robotistic
valuation of the segmentation network's output. Human common sense would valuate the confusion of the street with a pedestrian differently from the confusion event with the roles interchanged: an unjustified emergency brake is a much weaker consequence as potential harm than overlooking a person on the street and therefore should come with a significantly lower cost. While \commentRD{it seems reasonable to assume} that the confusion cost should be different from constant, it is \commentRD{ethically} much less evident, which numbers should \commentRD{explicitly} be used. \commentRD{In these situations of moral uncertainty, different ethical schools of thought may provide different answers}, with some refusing to weigh lives at all \cite{weighing}. In addition, legislation can put strong constraints on the choice as well. However, as the MAP principle and the ML decision rule \commentRD{already} define confusion cost matrices, \commentRD{choices about these numbers have already been made. We, therefore, aim to make more transparent the ethical dimension involved in making a choice regarding a decision rule with its corresponding cost matrix.} 

We realize that the ultimate step from probabilities to perception depends on cost matrices in a high dimensional value space $\mathcal{V}$ and that the selected valuation changes the perception. Thereby, it also changes the consequences, as, \emph{e.g.},~the precision and recall rates of specific classes. Furthermore, different cost matrices $C\in\mathcal{V}$ might express different ethical attitudes, like more egoistic (centred on the passenger in the (ego-) car) or altruistic (centred on public safety). Putting drivers first vs.~putting the public first has already been subject to intense public debate \cite{egocar}. 

In this paper, we do not intend to resolve the problem outlined above in any way. We present a numerical study that demonstrates the practical relevance of the problem by traveling through the value space within a triangle of robotistic and approximately egoistic and altruistic, respectively, cost value systems. Here the egoistic and altruistic cost matrices are set up in an \emph{ad hoc} manner and are not meant to \commentRD{accurately represent these attitudes}. Also, the matrices are by no means the most extreme ones spanning the value space. Nevertheless, when traveling through this small triangle in the large space of valuations $\mathcal{V}$, we see significant and relevant differences in the perception and measure consequences like the precision / recall and (segment-wise) false positive / negative rates for specific classes. 

The 
remainder of this paper is organized as follows: 
In \cref{sec:dr-nn} we describe \commentMR{our use-case for} decision rules in neural networks, in particular in semantic segmentation neural networks. Next, in \cref{sec:cost-based} we explain the concept of decision rules in general and how they can be modified by \commentMR{valuating} confusion costs between classes. We see various possibilities of defining the mentioned costs and provide two concrete examples in form of matrices in \cref{sec:setup}. Moreover, we present our spanned value space of confusion cost matrices and the setup for our experiments which follow in \cref{sec:experiments}. We show that different cost matrices are capable of considerably affecting the perception of a state-of-the-art semantic segmentation network in the setting of urban street scenes.

\section{Standard decision rule in neural networks} \label{sec:dr-nn}

Semantic segmentation is the task of assigning each pixel of an image to one of the predefined classes $\mathcal{K}=\{1,\ldots,N\}$. Suppose, we use a neural network for solving this task.
Let $x \in \{ (r,g,b) \} ^ {m \times n} ,\  (r,g,b) \in \{0,\ldots,255\}^3$ be an ``rgb'' (red, green and blue light additively colored) input image with resolution $m \times n$.
After processing the image $x$ with a neural network we obtain a posterior probability distribution $p_{ij}(k|x)$ over all classes $k\in\{1,\ldots\, N \}$ at location (pixel position in the image) $(i,j) \in \{1,\ldots,m\} \times \{1,\ldots,n\}$. The 3D tensor $p_{ij}(k|x)$ represents the softmax output of a neural network for semantic segmentation. The third dimension is given by the choice of $k\in\{1,\ldots\, N \}$.
This provided probability distribution expresses the confidence of the neural network as statistical prediction model to label the input correctly given the class $k$.
The pixel-wise classification is then performed by applying the $\mathop{argmax}$ function (pixel-wise) on the posterior probabilities / softmax output.
\commentMR{This kind of decision making is called} maximum a-posteriori probability (MAP) principle. 

In the field of Deep Learning, following the MAP as \textit{decision rule} is by far the most commonly used one. It maximizes the overall performance of a neural network, meaning in cases of large prediction uncertainty, this rule tends to predict classes that appear frequently in the dataset.
However, classes of potential high importance, like in autonomous driving \commentMR{the classes} traffic signs and humans, usually appear less frequently. These classes are rare in terms of the number of instances and the number of pixels in the dataset. This problem is in close connection to the fact that the MAP estimation considers all prediction mistakes to be equally serious \commentMR{which is in conflict with human intuition}.
Thus, a natural approach is to weight different prediction mistakes against each other.



\section{Cost-based decision rules in neural networks} \label{sec:cost-based}


Let $\Omega$ be a population consisting of $N \geq 2$ disjoint subsets. For each element $\omega \in \Omega$ we assume there exists one feature vector $x(\omega) \in S \subset \mathbb{R}^n$. Let
\begin{align}
    \phantom{\hat{k}}
    X&:\Omega \to S\ \\ 
    K&:\Omega \to \{1,\ldots,N\} = \mathcal{K} \phantom{\hat{k}}
\end{align}
be random variables for feature vector $x$ and class affiliation $k$, respectively.
A \textit{decision rule} can be defined as a map
\begin{align}
    d:\quad S\ &\to\ \mathcal{K} \\
    x(\omega) &\mapsto \hat{k}(\omega)
\end{align}
which assigns an element from the feature space to one class. We say, $d(x)=\hat{k}$ is the predicted class for feature vector $x$. Furthermore, we describe the a-posteriori probability of an object to belong to class $k$ given feature $x$ as
\begin{equation} \label{eq:apost}
p(k|x) := P(K=k\ |\ X=x).
\end{equation}
Usually, this probability is not known and needs to be estimated. We assume in the following that this is already accomplished, \emph{e.g.},~$p(k|x)$ is approximated by the softmax output of a neural network. 

Cost-based decision rules follow the idea of assigning one input to the class which minimizes the expected cost given one confusion cost function 
\begin{align*}
c: \mathcal{K} \times \mathcal{K} \to \mathbb{R}_{\geq0}:=[\ 0,\infty\ ).
\end{align*}
Considering all possible confusion cases we obtain a confusion cost matrix 
\begin{equation}
C := (c(\hat{k},k))_{\hat{k},k=1,\ldots,N} \in \mathcal{V} \subset \mathbb{R}_{\geq0}^{N \times N} \label{eq:conf-matrix}
\end{equation}
with $\hat{k}$ being the predicted class while $k$ being the target class and 
\begin{align}
    \mathcal{V} := \{\ C \in \mathbb{R}^{N \times N}\ |\ C_{jj} = 0, C_{ij} > 0, i,j \in \mathcal{K}\ \}
\end{align}
being the value space of all valid matrices $C$ for cost-based decision rules. Hence, all elements of a valid matrix must be positive except the diagonal elements, which must equal $0$, according to $\mathcal{V}$.
Strictly speaking, $\mathcal{V}$ consists of equivalence classes since each $C$ in combination with cost-based decision rules will produce the same output as $\mu C, \mu>0$, \emph{i.e.},~different scales of $C$ do not change the output. Therefore, rather the costs of the classes relative to each other are decisive for the output instead of the absolute values.

In order to understand the just stated fact we define the expected cost with respect to confusion cost functions via
\begin{align}
    \mathbb{E}[\ c(k^\prime,K)\ |\ X=x\ ]  = \sum_{k=1}^N c(k^\prime,k)\ p(k|x) \label{eq:exp-cost-sum} 
\end{align}
and the corresponding \textit{cost-based} decision rule as 
\begin{align}
d(x;C) :=& \argmin{k^\prime\in\{1,\ldots,N\}} \sum_{k=1}^N c(k^\prime,k)\ p(k|x) \label{eq:min-cost}\\ 
\stackrel{(\ref{eq:conf-matrix})}=& \argmin{k^\prime\in\{1,\ldots,N\}} C_{k^\prime} \cdot \vec{p}(x) = \hat{k} \phantom{\sum_{k^\prime}^N}
\end{align}
with $C_k := (C_{k1},\ldots,C_{kN})$ being the $k$-th row vector of $C\in\mathcal{V}$ and $\vec{p}(x):=(p(1|x),\ldots,p(N|x))^T$ being the posterior probabilities vector conditioned on \commentMR{the} feature $x$. This rule is optimal considering the expected costs.

Cost-based decision rules are strongly related to probability thresholding. The aim of probability thresholding is to make class predictions cost-sensitive during inference by moving the output threshold \commentMR{towards} inexpensive classes. This is achieved by defining a confusion cost function of the form
\begin{equation} \label{eq:cost-func}
c\left(\hat{k},k\right) := 
\begin{cases}
0 &,\ \text{if}\quad \hat{k}=k \\
\psi(k) &,\ \text{if}\quad \hat{k} \neq k
\end{cases}\ ,\
\psi(k) \in \mathbb{R}_{\geq0}
\end{equation}
with $\psi(k)>\psi(k^\prime)$ if we want the network to prefer predicting class $k$ to predicting class $k^\prime$. One special type of $c$ is the simple symmetric cost function \cite{Fahrmeir-1996}
\begin{equation} \label{eq:sym-cost}
c_s\left(\hat{k},k\right) := 
\begin{cases}
0 &,\ \text{if}\quad \hat{k}=k \\
\lambda &,\ \text{if}\quad \hat{k} \neq k
\end{cases}\ ,\
\lambda \in \mathbb{R}_{\geq0}
\end{equation}
whose incorporation in the cost-based decision rule is equivalent to the MAP principle. Given $c_s(\hat{k},k)$ all elements in the confusion cost matrix $C_s$ \commentMR{are} equal \commentMR{to} the constant $\lambda$ except the diagonal elements which \commentMR{are} equal $0$. Accordingly, the cost-based decision rule takes 
the form:
\begin{align}
    d(x;C_s)
    \stackrel{(\ref{eq:sym-cost})} =& \argmin{k\in\{1,\ldots,N\}} \sum_{k^\prime = 1, k^\prime \neq k}^N \lambda \cdot p(k^\prime|x) \\
    =& \argmin{k\in\{1,\ldots,N\}} 1 - p(k|x) \phantom{\sum_{k^\prime}^N} \\
    \phantom{\sum_{k^\prime}^N} =& \argmax{k\in\{1,\ldots,N\}}p(k|x)\ =:\ d_\textit{Bayes}(x). \label{eq:bayes-rule}
\end{align}
In decision theory \cref{eq:bayes-rule} is the definition of the \textit{Bayes} decision rule which is equivalent to the MAP principle
and therefore also to the \commentMR{default} classification principle in neural networks. However, the simple symmetric cost function implies an equal class weighting, \emph{i.e.},~weighting every confusion between two classes (or each type of misclassification) equally. Depending on the purpose, this setting does not reflect the intuition of most people but is still applied in most deep learning state-of-the-art models.

\begin{figure}
\begin{tikzpicture}
\node [align=center] at (.25\textwidth,0) {\includegraphics[width=.21\textwidth]{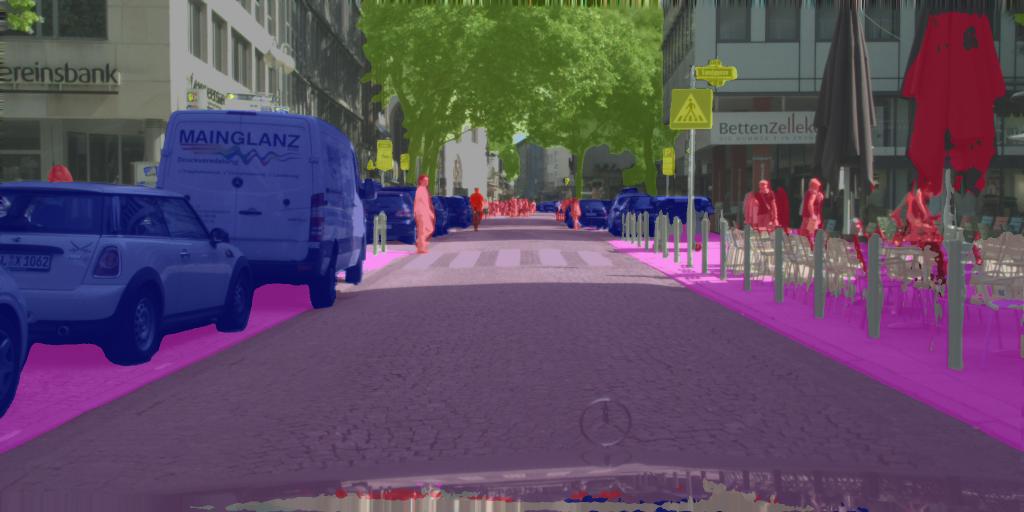}};
\node [align=center] at (.5\textwidth,0) {\includegraphics[width=.21\textwidth]{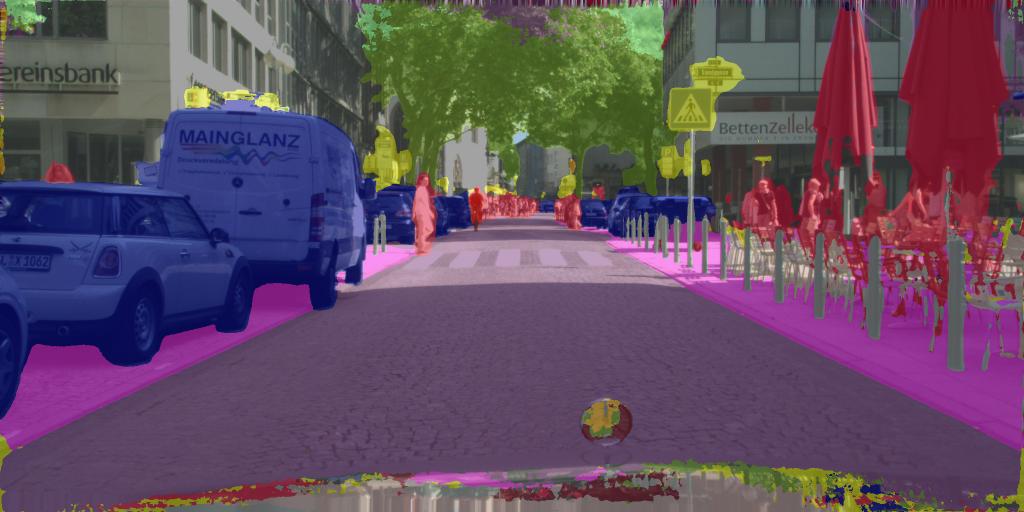}};
\node [align=center] at (.25\textwidth,1.4) {Bayes};
\node [align=center] at (.5\textwidth,1.4) {\phantom{y}Maximum Likelihood\phantom{y}};
\end{tikzpicture}
\caption{Illustration of two segmentation masks obtained with the Bayes decision rule (right) and the Maximum Likelihood decision rule (left). The difference between these two masks lies in the adjustment with the (pixel-wise) prior class probabilities in the decision rule during inference.}
\end{figure}

A mathematically natural way to approach this problem is exchanging the simple symmetric with the inverse proportional cost function \cite{Fahrmeir-1996} which is another special type of $c$.
In light of confusion costs the latter cost function
\begin{equation} \label{eq:inv-cost}
c_p\left(\hat{k},k\right) := 
\begin{cases}
0 &,\ \text{if}\quad \hat{k}=k \\
\lambda/p(k) &,\ \text{if}\quad \hat{k} \neq k
\end{cases}\ ,\
\lambda \in \mathbb{R}_{\geq0}
\end{equation} 
weights \commentMR{each} confusion with the inverse prior probability $1/p(k), p(k)\in(0,1)$ of the potential target class $\commentMR{k}$. In neural networks the class appearance frequencies in the training data correspond approximately to the priors. Considering the priors, we can put more emphasis on finding classes which are rare, \emph{i.e.}, classes which have a low prior probability.
The decision rule resulting from this is the \textit{Maximum Likelihood} (ML) decision rule
\begin{equation} \label{eq: ml-rule}
d_{\textit{ML}}(x):=\argmax{k\in\{1,\ldots,N\}}p(x|k).
\end{equation}
Now $x$ is mapped to the class $k$ for which the observed features are most typical\commentMR{,} independent of a prior belief about the class frequencies.
As presented in \cite{chan19}, with respect to rare classes the application of the ML 
rule significantly reduces the number of false negative \commentMR{(overlooked) segments for rare classes,} but to the detriment of producing substantially more false positive segment predictions. One might argue that there is a ``sweet spot'' where the two error rates, the positive and negative \commentMR{one}, are optimal. However, one might also argue that certain classes are still underweighted relative to others. We address both problems by applying the cost-based decision rule in combination with adjusting the confusion cost matrix $C$.


\section{Setup of experiments} \label{sec:setup}

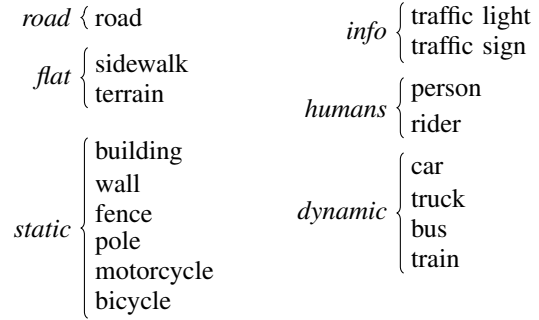
\begin{figure}
    \centering
    \begin{tikzpicture}

\def\aw{0}
\def\bw{0}
\def\cw{0}
\def\dw{3}
\def\ew{3}
\def\fw{3}

\def\ah{0}
\def\bh{-0.6}
\def\ch{-1.8}
\def\dh{0}
\def\eh{-1.0}
\def\fh{-2.0}

\node[text width=4cm] at (\aw,\ah) {road};
\draw[decorate, decoration={brace,mirror}]  ({\aw-2.1},{\ah+0.15}) -- node[left=0.6ex] {\textit{road}}  ({\aw-2.1},{\ah-0.15});

\node[text width=4cm] at (\bw,\bh) {sidewalk};
\node[text width=4cm] at (\bw,{\bh-0.4}) {terrain};
\draw[decorate, decoration={brace,mirror}]  ({\bw-2.1},{\bh+0.2}) -- node[left=0.6ex] {\textit{flat}}  ({\bw-2.1},{\bh-0.6});

\node[text width=4cm] at (\cw,\ch) {building};
\node[text width=4cm] at (\cw,{\ch-0.4}) {wall};
\node[text width=4cm] at (\cw,{\ch-0.8}) {fence};
\node[text width=4cm] at (\cw,{\ch-1.2}) {pole};
\node[text width=4cm] at (\cw,{\ch-1.6}) {motorcycle};
\node[text width=4cm] at (\cw,{\ch-2.0}) {bicycle};
\draw[decorate, decoration={brace,mirror}]  ({\cw-2.1},{\ch+0.2}) -- node[left=0.6ex] {\textit{static}}  ({\cw-2.1},{\ch-2.2});

\node[text width=1.6cm] at (\dw,\dh) {traffic light};
\node[text width=1.6cm] at (\dw,{\dh-0.4}) {traffic sign};
\draw[decorate, decoration={brace,mirror}]  ({\dw-0.9},{\dh+0.2}) -- node[left=0.6ex] {\textit{info}}  ({\dw-0.9},{\dh-0.6});

\node[text width=1.6cm] at (\ew,\eh) {person};
\node[text width=1.6cm] at (\ew,{\eh-0.4}) {rider};
\draw[decorate, decoration={brace,mirror}]  ({\ew-0.9},{\eh+0.2}) -- node[left=0.6ex] {\textit{humans}}  ({\ew-0.9},{\eh-0.6});

\node[text width=1.6cm] at (\fw,\fh) {car};
\node[text width=1.6cm] at (\fw,{\fh-0.4}) {truck};
\node[text width=1.6cm] at (\fw,{\fh-0.8}) {bus};
\node[text width=1.6cm] at (\fw,{\fh-1.2}) {train};
\draw[decorate, decoration={brace,mirror}]  ({\fw-0.9},{\fh+0.2}) -- node[left=0.6ex] {\textit{dynamic}}  ({\fw-0.9},{\fh-1.4});

\end{tikzpicture}
    \caption{Class aggregates of Cityscapes classes that we use for simplicity in our experiments. Note that in the Cityscapes labeling motorcycles and bicycles in motion adhere to the class ``rider''.  }
    \label{fig:mapping}
\end{figure}

For our experiments we use the Cityscapes dataset with $19$ semantic classes. In order to reduce the number of confusion cost values to be specified for the matrix $C$ we \commentMR{aggregate} classes that are treated similarly considering confusion costs, see \cref{fig:mapping} for a first attempt although refined aggregations are probably more appropriate. 

With $6$ aggregated classes we define a $6\times6$ matrix. For performance evaluation we map the reduced matrix back to full $19\times19$ size such that all combinations between classes out of two aggregates have an equal confusion cost, 
\emph{i.e.}, for two different non-empty aggregates $\mathcal{I}, \mathcal{J} \subset \mathcal{K}$ it holds
\begin{align}
    \mathcal{I} \cap \mathcal{J} &= \emptyset\  \\
    \Leftrightarrow c(i,j) &= c(i^\prime,j^\prime)\ \forall\ i,i^\prime \in \mathcal{I},\ j,j^\prime \in \mathcal{J}.
\end{align}
In addition, we set a small $\epsilon=0.1$ for all confusions between different classes within an aggregate so that we apply the Bayes decision rule (only within an aggregate) without affecting the cost-based \commentMR{decision}
between \commentMR{aggregated} classes,
\emph{i.e.}, for each non-empty aggregate $\mathcal{I} \in \mathcal{K}$ it holds
\begin{align}
    c(i,i^\prime) &= \epsilon\ \forall\ i \neq i^\prime \in \mathcal{I} \\
    c(i,i) &= 0\ \forall\ i \in \mathcal{I}.
\end{align}
Note that we suppress the ``sky'' class in our class \commentMR{aggregation} although it is one of the originally trained classes. The reason is that we believe \commentMR{that} overlooking the sky does not result in dangerous traffic scenarios. Therefore, we prevent the network from predicting \commentMR{sky} by setting $C_\textit{sky}^T= \{ M \}^N$ with $M=1000$ being a sufficiently large cost value.
This implies that the confusion of any (target) class with sky is valuated with high cost. We set the cost for the converse confusion, when sky is the target class, to a constant value in order to not affect the class prediction between the remaining classes.

\begin{figure}
    \centering
    \begin{tikzpicture}
    \node [align=center] at (0,0) {\includegraphics[width=.3\textwidth]{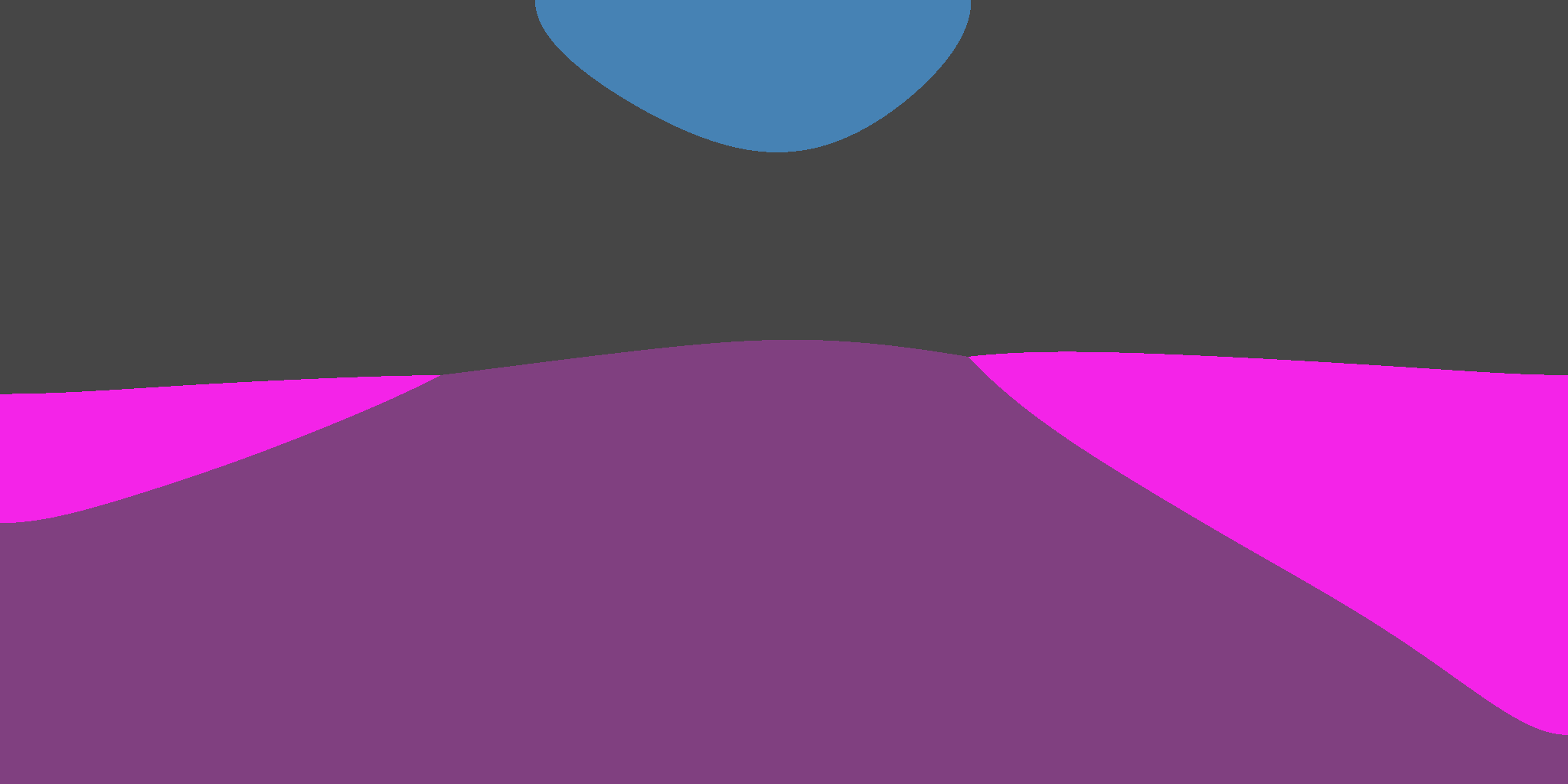}};
    \node [align=center] at (4,0) {\includegraphics[width=.08\textwidth]{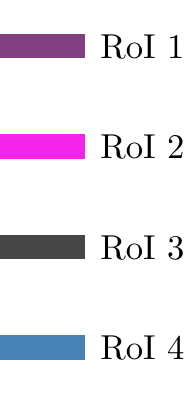}};
    \end{tikzpicture}
    \caption{Regions of interest derived from the priors of the classes building, road, sidewalk and sky in the Cityscapes dataset. 
    }
    \label{fig:roi}
\end{figure}

To gain further insight we define image regions of interest (RoI). These regions are derived from the pixel-wise class frequencies (priors) of the classes ``road'', ``sidewalk'', ``building''  and ``sky'' in the Cityscapes dataset. We obtain the $4$ regions of interest (or $5$ regions as the sidewalk RoI consists of two connected components) by assigning each pixel to the class with the highest class appearance frequency at the corresponding pixel location, see \cref{fig:roi}.

\begin{figure}
    \centering
    \input{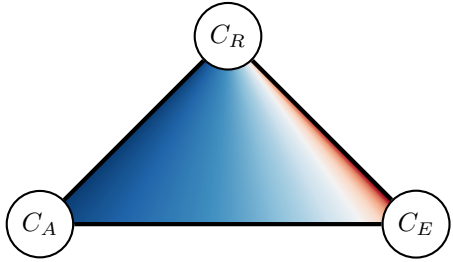}
    \caption{Confusion cost matrix space $V$ spanned by our exemplary altruistic ($C_A$) and egoistic ($C_E$) cost matrix and the robotistic ($C_R$) cost matrix. Inside the triangle as heatmap the behavior of $\textit{rec}(\ V(C)\ |\ \textit{person}\ )$, the recall of person pixels. Blue indicates high recall, red indicates low recall.}
    \label{fig:recall-person}
\end{figure}

\begin{figure}
    \centering
    \input{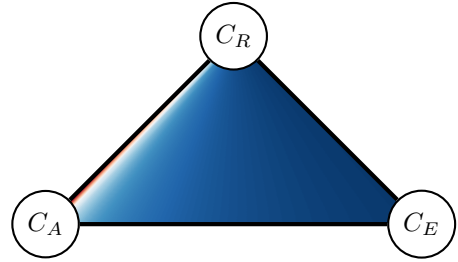}
    \caption{Confusion cost matrix space $V$ spanned by our 
    \commentMR{exemplary} altruistic ($C_A$) and egoistic ($C_E$) cost matrix and the robotistic ($C_R$) cost matrix. Inside the triangle as heatmap the behavior of $\textit{rec}(\ V(C)\ |\ \textit{building}\ )$, the recall of building pixels. Blue indicates high recall, red indicates low recall.}
    \label{fig:recall-buidling}
\end{figure}

For our experiments we further define two confusion cost matrices representing two extreme views in traffic scenes. On the one hand, we define the ``altruistic'' \commentMR{matrix} $C_A$ that prioritizes all traffic participants and particularly humans. On the other hand, we define the ``egoistic'' \commentMR{matrix} $C_E$ that only prioritizes the safety and comfort of the passenger inside the (ego-) car. The chosen cost values can be viewed in \cref{fig:cost-mat}. We compare the corresponding predictions with each other and also with the Bayes rule's prediction, respectively. The Bayes decision rule implies the \commentMR{matrix} 
$C_R := (c_s(\hat{k},k))_{\hat{k},k = 1,\ldots,N}$ which we \commentMR{term} in the following \commentMR{the} ``robotistic'' confusion cost matrix. 
This method is robotistic in the sense that, in any event, the only goal is to minimize all error rates.
The convex combinations of these three presented matrices span a confusion value space 
\begin{align}
\begin{split}
    V:=\{\ C \in \mathcal{V}\ |\ &\alpha C_R + \beta C_A + \gamma C_E = C, \\  
            &\alpha + \beta + \gamma = 1,\ \alpha,\beta,\gamma \geq 0 \ \}
\end{split}
\end{align}
(see \cref{fig:recall-person} and \cref{fig:recall-buidling}). It is important to emphasize that $V \subset \mathcal{V}$ is only one subspace of a far bigger possible value space. There are even more extreme cost matrices that enlarge the space dramatically. There are also cost matrices expressing views in a completely different direction and therefore increasing the dimensionality of the space. However, our presented $V$ is sufficient in order to show that it is already capable of changing our model's perception significantly.

\begin{table}
\begin{center}
\resizebox{0.85\columnwidth}{!}{%
\begin{tabular}{|l|cc|cc|}
\hline
Cost matrix & Class & RoI & Precision & Recall \\
\hline
Altruistic  & Person    & 1   & $41.12\%$ & $\mathbf{99.81}\%$ \\
Robotistic  & Person    & 1   & $89.87\%$ & $94.98\%$ \\
Egoistic    & Person    & 1   & $\mathbf{93.88}\%$ & $70.07\%$ \\
\hline
Altruistic  & Person    & 2   & $39.42\%$ & $\mathbf{99.86}\%$ \\
Robotistic  & Person    & 2   & $88.36\%$ & $93.93\%$ \\
Egoistic    & Person    & 2   & $\mathbf{95.07}\%$ & $54.81\%$ \\
\hline
Altruistic  & Building  & 1   & $22.56\%$ & $93.65\%$ \\
Robotistic  & Building  & 1   & $\mathbf{80.99}\%$ & $94.94\%$ \\
Egoistic    & Building  & 1   & $15.15\%$ & $\mathbf{99.93}\%$ \\
\hline
Altruistic  & Building  & 2   & $24.94\%$ & $95.22\%$ \\
Robotistic  & Building  & 2   & $\mathbf{87.76}\%$ & $94.58\%$ \\
Egoistic    & Building  & 2   & $18.48\%$ & $\mathbf{99.90}\%$ \\
\hline
\end{tabular}%
}
\end{center}
\caption{\commentRC{Precision and recall \commentMR{rates for} the three different cost matrices. \commentMR{The rates are computed for the classes person and building in the street and the sidewalk RoIs, \emph{i.e.}, RoI $1$ and $2$. 
}}}
\label{fig:results-prc-rec}
\end{table}

\section{Experiments} \label{sec:experiments}

\begin{figure*}
\begin{tikzpicture}
\node [align=left] at (.0\textwidth,0) {\includegraphics[width=.325\textwidth]{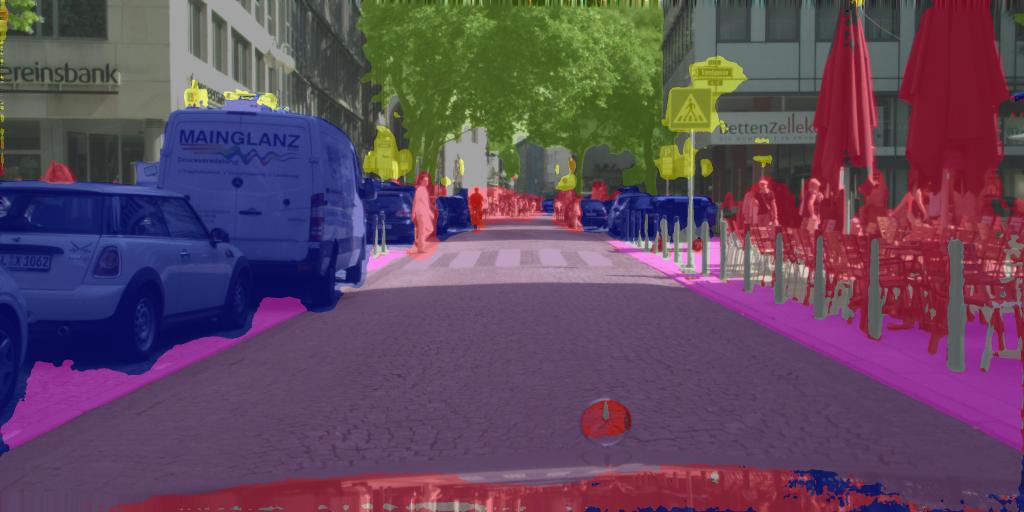}};
\node [align=left] at (.33\textwidth,0) {\includegraphics[width=.325\textwidth]{graphics/bayes76.jpg}};
\node [align=left] at (.66\textwidth,0) {\includegraphics[width=.325\textwidth]{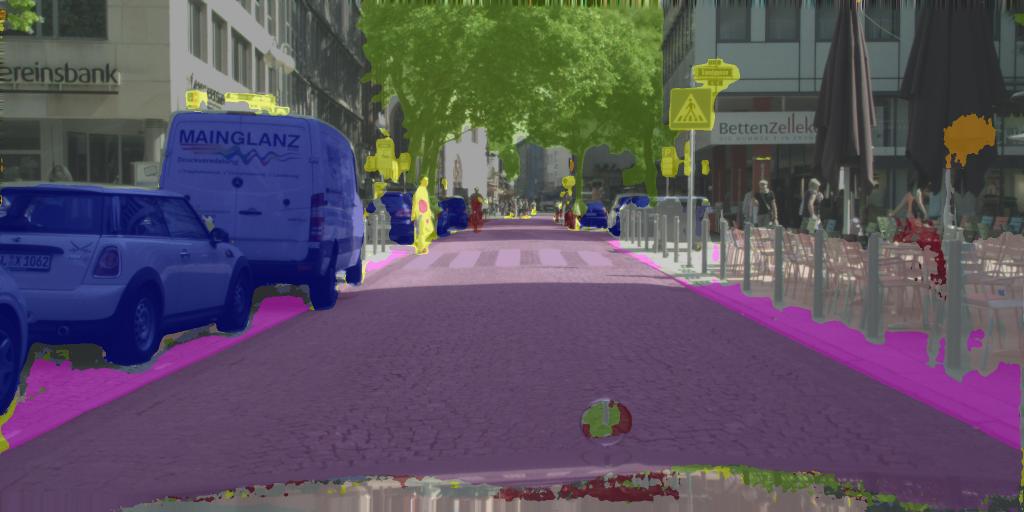}};
\node [align=center] at (0\textwidth,1.75) {Altruistic};
\node [align=center] at (.33\textwidth,1.75) {Robotistic};
\node [align=center] at (.66\textwidth,1.75) {Egoistic};
\end{tikzpicture}
\caption{Illustration of three semantic segmentation masks and different perception obtained by the application of cost-based decision rules with an altruistic, a simple symmetric (robotistic) and an egoistic cost matrix.}
\end{figure*}
\begin{figure*}
    \vspace{-0.25cm}
    \centering
    \begin{tikzpicture}

\def\w{-12}
\def\wp{-11.55}


\node [align=left,rotate=45,text width=1.65cm] at ({\wp+1},2) {$\mathrm{``road"}$};
\node [align=left,rotate=45,text width=1.65cm] at ({\wp+1.86},2.) {$\mathrm{``flat"}$};
\node [align=left,rotate=45,text width=1.65cm] at ({\wp+2.72},2.) {$\mathrm{``static"}$};
\node [align=left,rotate=45,text width=1.65cm] at ({\wp+3.58},2.) {$\mathrm{``info"}$};
\node [align=left,rotate=45,text width=1.65cm] at ({\wp+4.44},2.) {$\mathrm{``human"}$};
\node [align=left,rotate=45,text width=1.65cm] at ({\wp+5.35},2.) {$\mathrm{``dynamic"}$};

\node [align=left,text width=1.65cm] at ({\w+6.8},1.12) {$\mathrm{``road"}$};
\node [align=left,text width=1.65cm] at ({\w+6.8},0.69) {$\mathrm{``flat"}$};
\node [align=left,text width=1.65cm] at ({\w+6.8},0.27) {$\mathrm{``static"}$};
\node [align=left,text width=1.65cm] at ({\w+6.8},-0.16) {$\mathrm{``info"}$};
\node [align=left,text width=1.65cm] at ({\w+6.8},-0.58) {$\mathrm{``human"}$};
\node [align=left,text width=1.65cm] at ({\w+6.8},-1.05) {$\mathrm{``dynamic"}$};

\node [align=center] at (\w,0) {$C_A=$};
\node [align=center] at ({\w+3.2},0) 
{$\begin{pmatrix}
  0 & 10^0 & 10^1 & 10^2 & 10^3 & 10^2 \\
10^0 &    0 & 10^1 & 10^2 & 10^3 & 10^2 \\
10^0 & 10^0 &    0 & 10^2 & 10^2 & 10^1 \\
10^0 & 10^0 & 10^0 &    0 & 10^3 & 10^2 \\
10^0 & 10^0 & 10^0 & 10^2 &    0 & 10^1 \\
10^0 & 10^0 & 10^0 & 10^2 & 10^3 &    0 \\
\end{pmatrix}$};

\node [align=center] at (-3.2,0) {$C_E=$};
\node [align=center] at (0,0) 
{$\begin{pmatrix}
   0 & 10^0 & 10^3 & 10^2 & 10^1 & 10^2 \\
10^0 &    0 & 10^3 & 10^2 & 10^1 & 10^2 \\
10^1 & 10^0 &    0 & 10^3 & 10^0 & 10^1 \\
10^1 & 10^0 & 10^3 &    0 & 10^0 & 10^1 \\
10^1 & 10^0 & 10^3 & 10^2 &    0 & 10^2 \\
10^1 & 10^1 & 10^3 & 10^2 & 10^2 &    0 \\
\end{pmatrix}$};

\draw[decorate, decoration={brace}]  (3,1.2) -- node[below=0.6ex, rotate=90] {\textit{prediction in rows}}  (3,-1.2);
\draw[decorate, decoration={brace}]  (-2.4,1.5) -- node[above=0.4ex] {\textit{potential target in columns}}  (2.3,1.5);

\end{tikzpicture}
    \caption{Two extreme confusion cost matrices that we study in our experiments. $C_A$ represents the altruistic view prioritizing all traffic participants and particularly pedestrians. $C_E$ represents the egoistic view prioritizing only the passenger in the (ego-) car. One element in the matrix expresses the cost that arises if we predict the class corresponding to the row and we confuse it with the potential target class corresponding to the column.}
    \label{fig:cost-mat}
\end{figure*}

As part of autonomous car driving systems, interpreting visual inputs is crucial in order to obtain a full understanding of the car's environment. 
The inference of an image in semantic segmentation \cite{cityscapes16,Everingham2015} is performed at pixel level combining object detection and localization. In recent years, deep learning has achieved great success in a wide range of problems including semantic segmentation. Most state-of-the-art models are built on deep convolutional neural networks (CNNs) \cite{Krizhevsky2012,Simonyan14}. 
One important contribution to CNNs for semantic segmentation is the Fully Convolutional Network (FCN) \cite{Long16} which introduces end-to-end training taking input of arbitrary size and producing output of equal size. The network is one of the first using an encoder-decoder structure \cite{Badrinarayanan15,RonnebergerFB15} whose encoder part is a classification network followed by the decoder part that projects convolved learned features back onto full pixel resolution. With the integration of atrous (also called dilated) convolutions \cite{Yu15}, that allows an exponential increase of the network's receptive field without loss of resolution, the performance of semantic segmentation networks is further significantly improved. One advanced module based on the latter operation is atrous spatial pyramid pooling (ASPP) \cite{chen16}. It is one of the main contributions to the network DeepLabv3+ \cite{chen18} which we use in the following in our experiments.


We demonstrate the performance of cost-based decision rules with different confusion cost matrices on the Cityscapes \cite{cityscapes16} validation dataset. DeepLabv3+ is already pretrained on the latter dataset and implemented in TensorFlow \cite{tensorflow2015-whitepaper}. The implementation and tuned weights are publicly available on GitHub. As network backbone, we choose the modified version of the Xception model \cite{Chollet16} that attains an mIoU score of $79.55 \%$ on the Cityscapes validation set with the application of the MAP / Bayes decision rule.

In the following, we perform our analysis for the classes ``person'' and ``building'' which are key classes in our problem setting of autonomous driving for the altruistic and egoistic view, respectively. Furthermore, we focus our studies on the regions of interest 1 \& 2, the near field perception in front of the (ego-) car and to the side of the (ego-) car. 

\paragraph{Pixel-wise precision vs. recall.}
\commentRC{
For evaluation we first consider precision and recall. These two metrics are closely connected to the quantities false positive and false negative pixel predictions. A predicted pixel is a false positive (FP) if it falsely indicates an object's presence. A predicted pixel ignoring the presence of a present object is a false negative (FN).
Therefore, precision is the percentage of a model's predicted pixels that match the ground truth, while recall is the percentage of ground truth pixels that a model predicts correctly, \emph{i.e.},
\begin{align}
    \textit{prc} &= TP\ /\ (TP+FP)\\
    \textit{rec} &= TP\ /\ (TP+FN)
\end{align}
with $TP$ being the true positives (pixels correctly classified according to the ground truth).
The two evaluation metrics can be formulated as maps
\begin{align}
    \textit{prc,rec}: V &\to [\ 0,1\ ]
\end{align}
expressing the neural network's predictive power 
depending on $C \in V$. The higher the value, the less prediction mistakes we obtain regarding falsely detected and non-detected pixels, respectively.
}
The precision and recall scores 
of the different cost matrices in different regions of interest
can be found in \cref{fig:results-prc-rec}. 

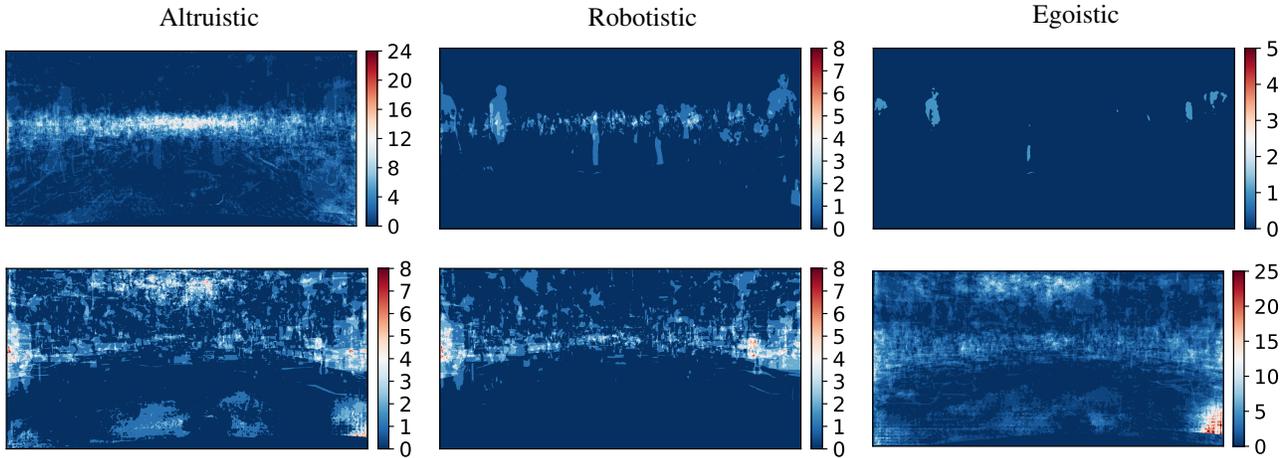
\begin{figure*}
\begin{tikzpicture}
\node [align=left] at (.0\textwidth,0) {\includegraphics[width=.325\textwidth]{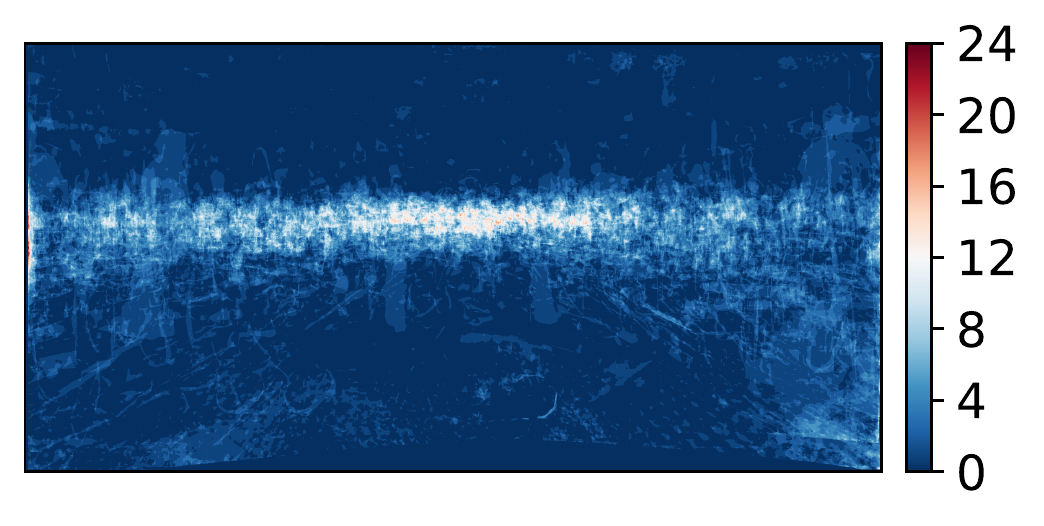}};
\node [align=left] at (.33\textwidth,0) {\includegraphics[width=.325\textwidth]{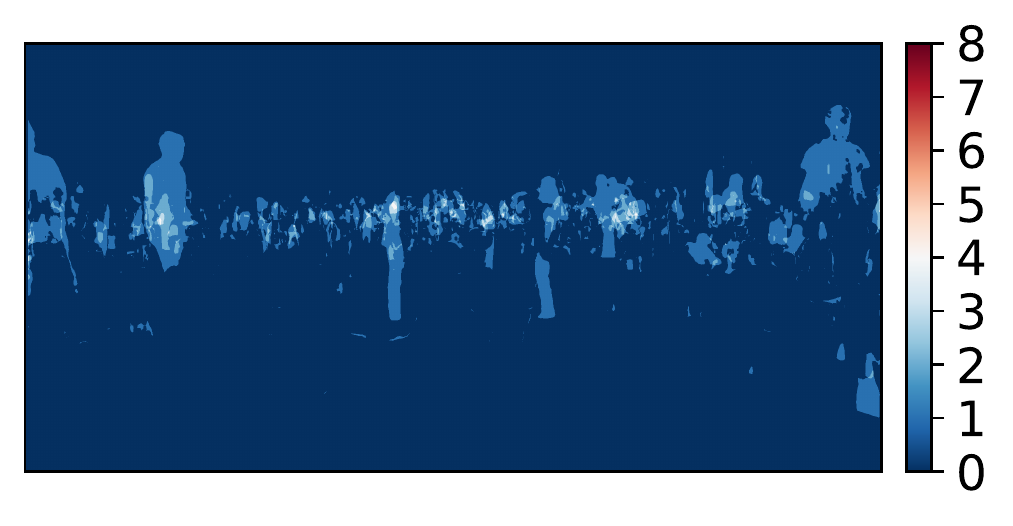}};
\node [align=left] at (.66\textwidth,0) {\includegraphics[width=.325\textwidth]{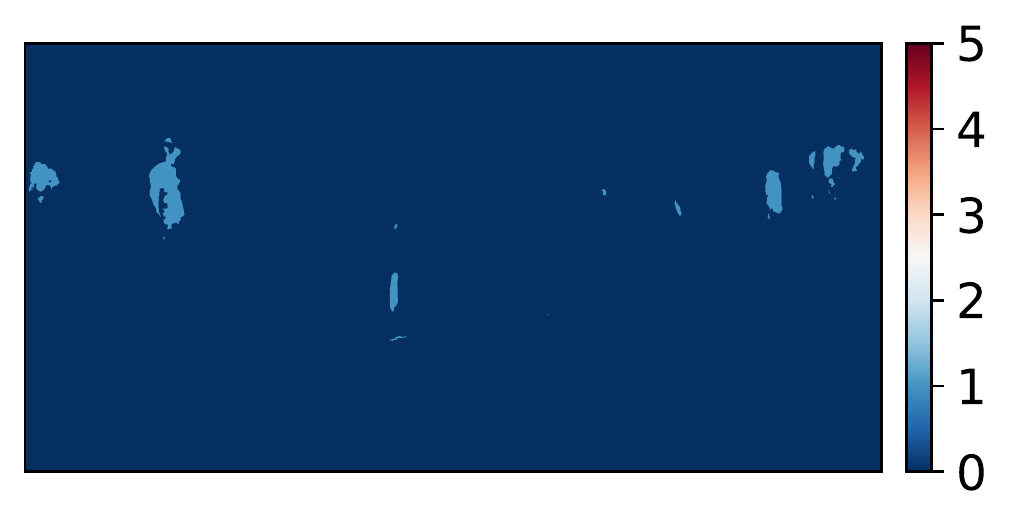}};
\node [align=center] at (0\textwidth,1.65) {Altruistic};
\node [align=center] at (.33\textwidth,1.65) {Robotistic};
\node [align=center] at (.66\textwidth,1.65) {Egoistic};

\node [align=left] at (.0\textwidth,-2.93) {\includegraphics[width=.325\textwidth]{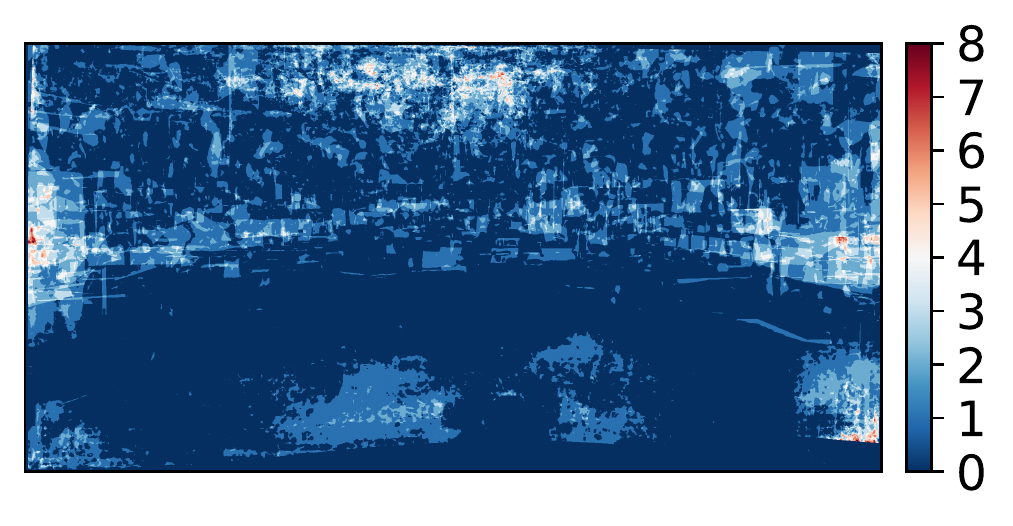}};
\node [align=left] at (.33\textwidth,-2.93) {\includegraphics[width=.325\textwidth]{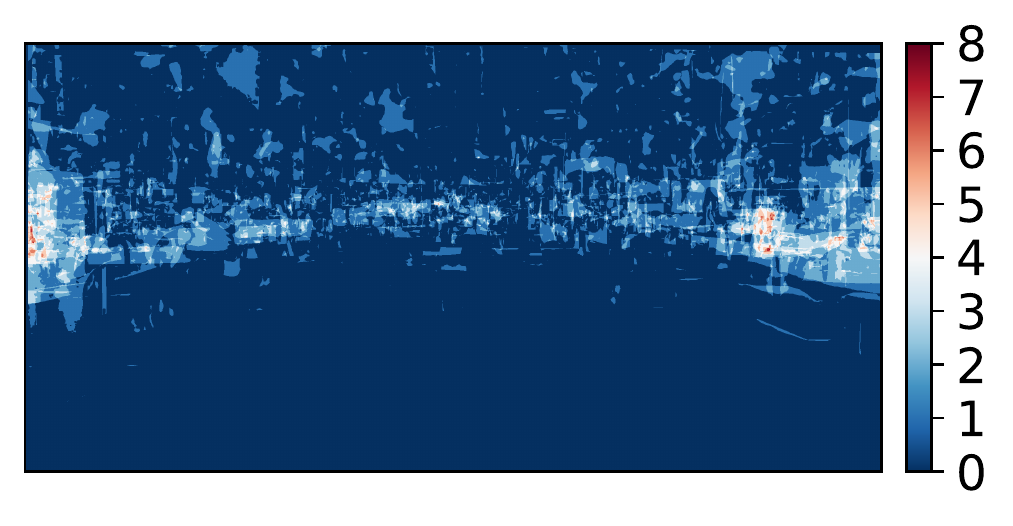}};
\node [align=left] at (.66\textwidth,-2.93) {\includegraphics[width=.325\textwidth]{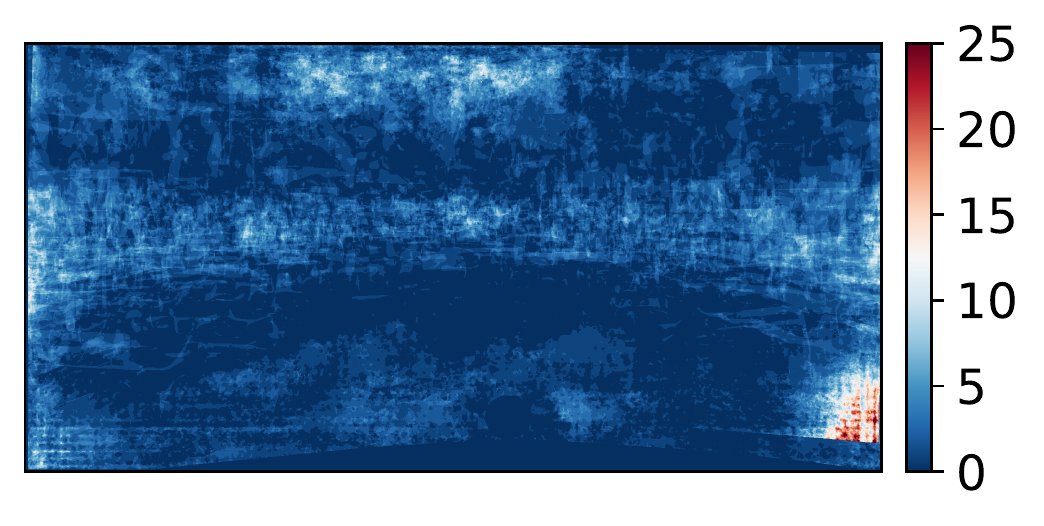}};
\end{tikzpicture}
\caption{Falsely detected (false positive) person (top row) and building (bottom row) segments.}
\label{fig:fp}
\end{figure*}

For the class person we observe that the recall is maximized when using $C_A$. Compared to $C_R$ the reduction is $4.83$ percent points in the \commentMR{street} RoI and even $5.93$ percent points in the sidewalk RoI. Even if the recall of person instances is already impressively high, $C_A$ is still capable of boosting the performance in this metric such that nearly no person pixels are missed. However, to a striking detriment, the precision is reduced by about $48$ percent points in both RoIs down to $41.12\%$ and $39.42\%$, respectively.
When using $C_E$ persons are ignored to a large extent leading to a recall reduction of $24.91$ percent points in the frontal RoI and $39.12$ percent points in the sidewalk RoI in comparison to $C_R$. Consequently, the precision is increased by $4.01$ and $6.71$ percent points, respectively. With $C_E$ DeepLabv3+ only predicts persons if the network indicates a high \commentMR{confidence} about its decision. As expected there is a trade off between the metrics, \emph{i.e.}, increasing one performance measure decreases the other and vice versa. Also noteworthy from this analysis is that DeepLabv3+ confuses only persons which are not completely visible, \emph{e.g.},~persons standing behind cars or around corners. Only small parts of person instances are mainly overlooked, see also \cref{fig:fp}.

For the class building we also observe this trade off but only between $C_E$ and $C_R$. $C_E$ improves the recall by $4.99$ and $5.32$ percent points while reducing the precision \commentMR{substantially by} $65.84$ and $69.28$ percent points, respectively, for the \commentMR{street} and the sidewalk RoI.

\begin{figure*}
\begin{tikzpicture}
\node [align=left] at (.0\textwidth,0) {\includegraphics[width=.325\textwidth]{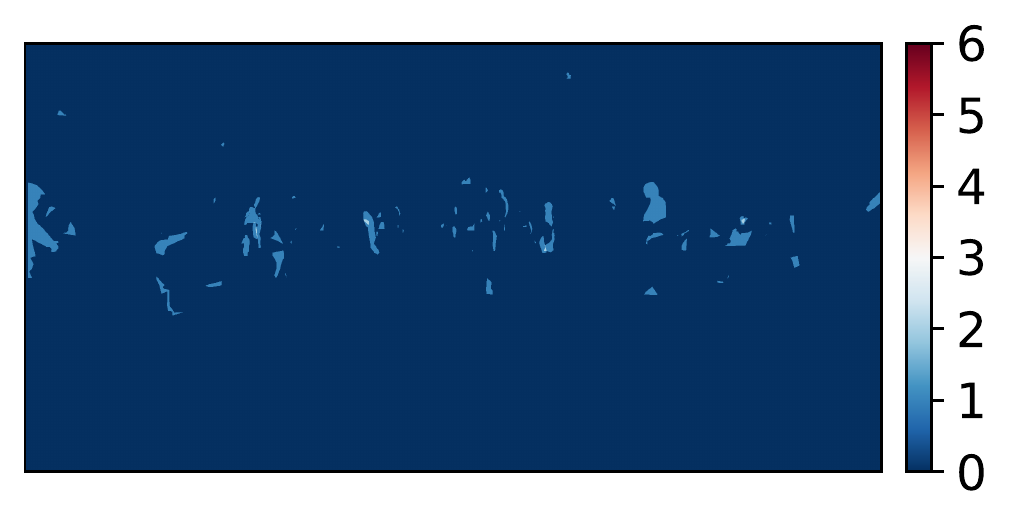}};
\node [align=left] at (.33\textwidth,0) {\includegraphics[width=.325\textwidth]{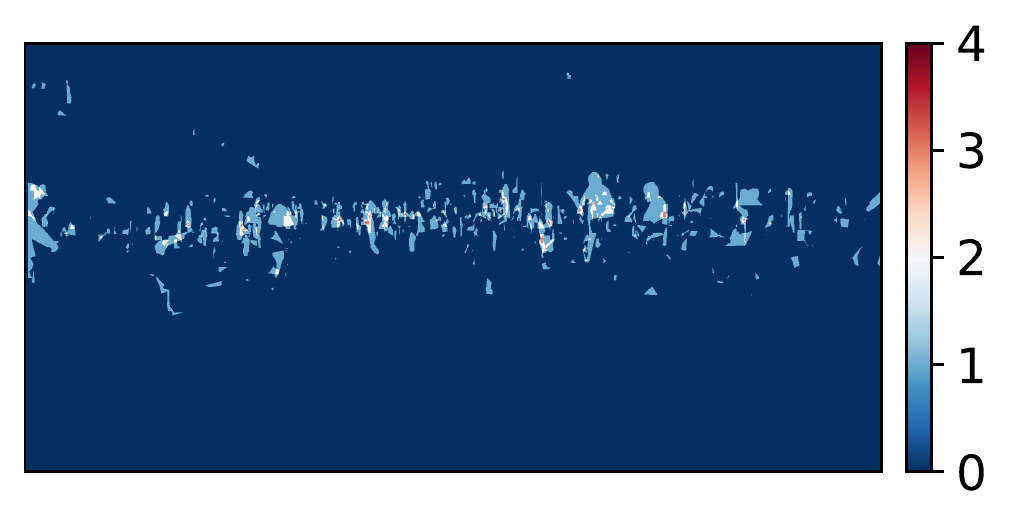}};
\node [align=left] at (.66\textwidth,0) {\includegraphics[width=.325\textwidth]{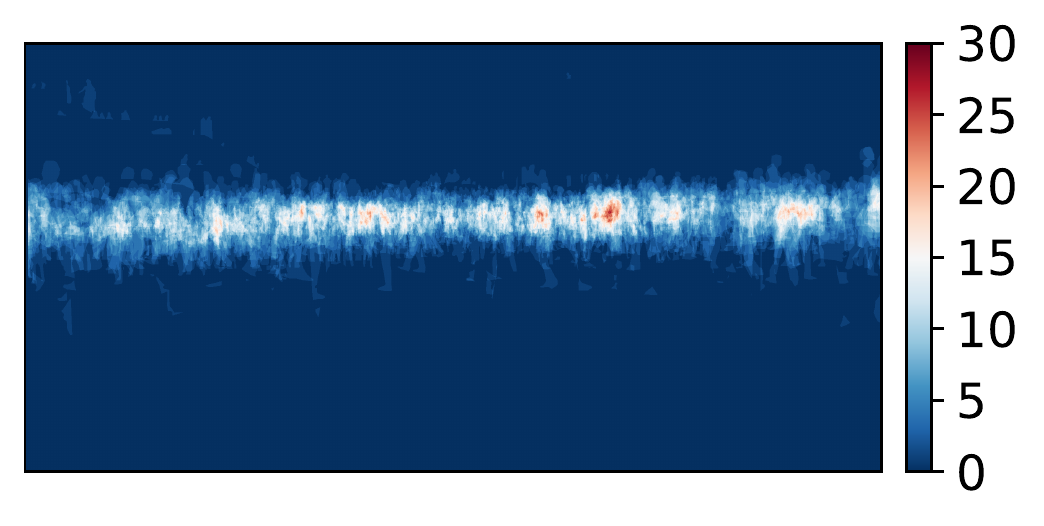}};
\node [align=center] at (0\textwidth,1.65) {Altruistic};
\node [align=center] at (.33\textwidth,1.65) {Robotistic};
\node [align=center] at (.66\textwidth,1.65) {Egoistic};

\node [align=left] at (.0\textwidth,-2.93) {\includegraphics[width=.325\textwidth]{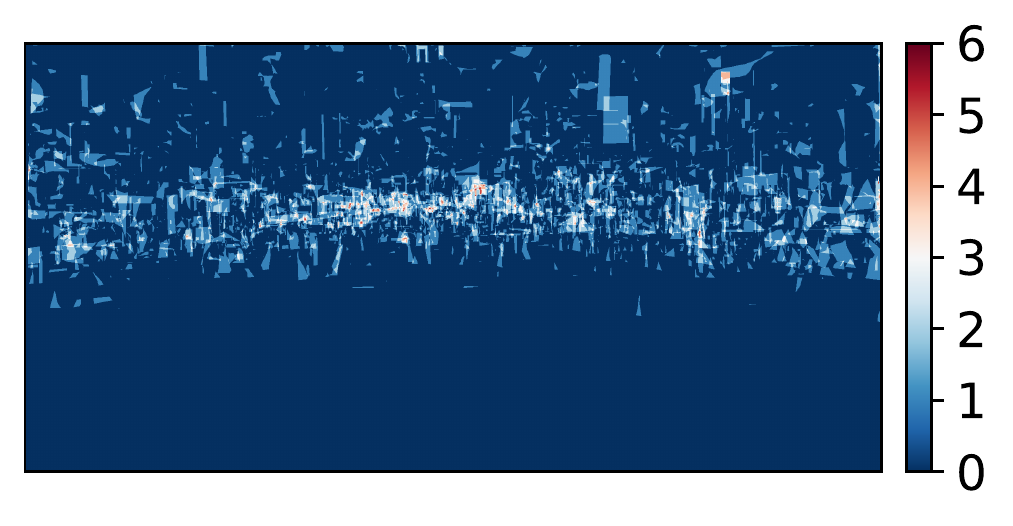}};
\node [align=left] at (.33\textwidth,-2.93) {\includegraphics[width=.325\textwidth]{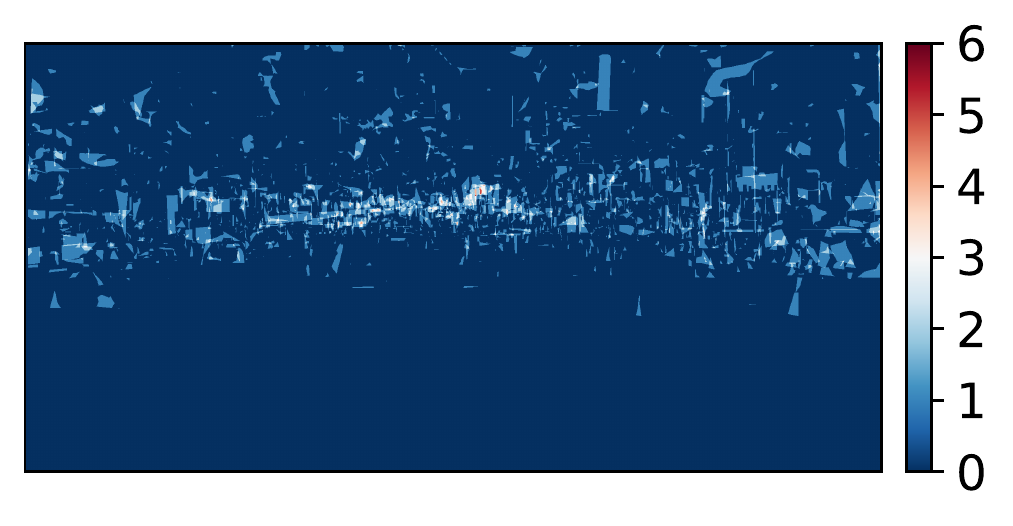}};
\node [align=left] at (.66\textwidth,-2.93) {\includegraphics[width=.325\textwidth]{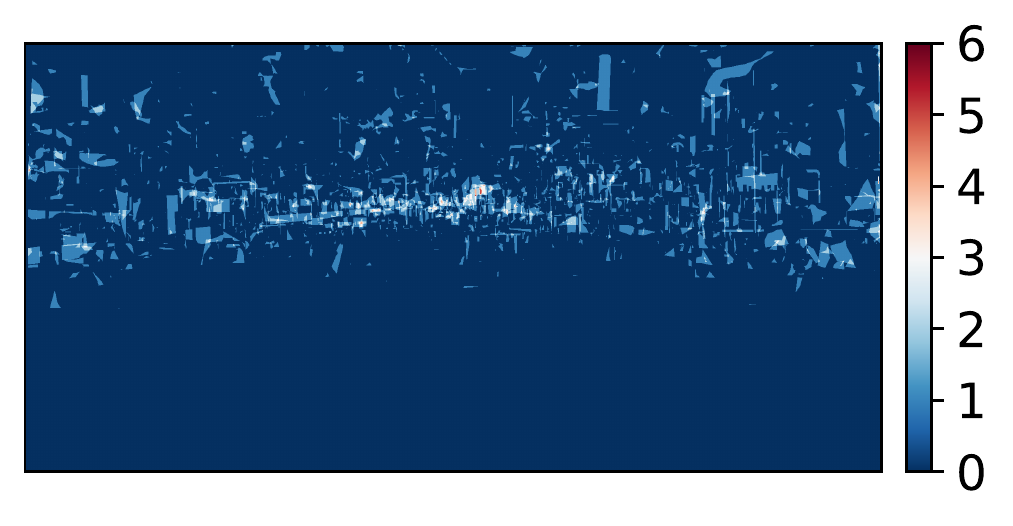}};
\end{tikzpicture}
\caption{Non-detected (false negative) person (top row) and building (bottom row) segments.}
\label{fig:fn}
\end{figure*}
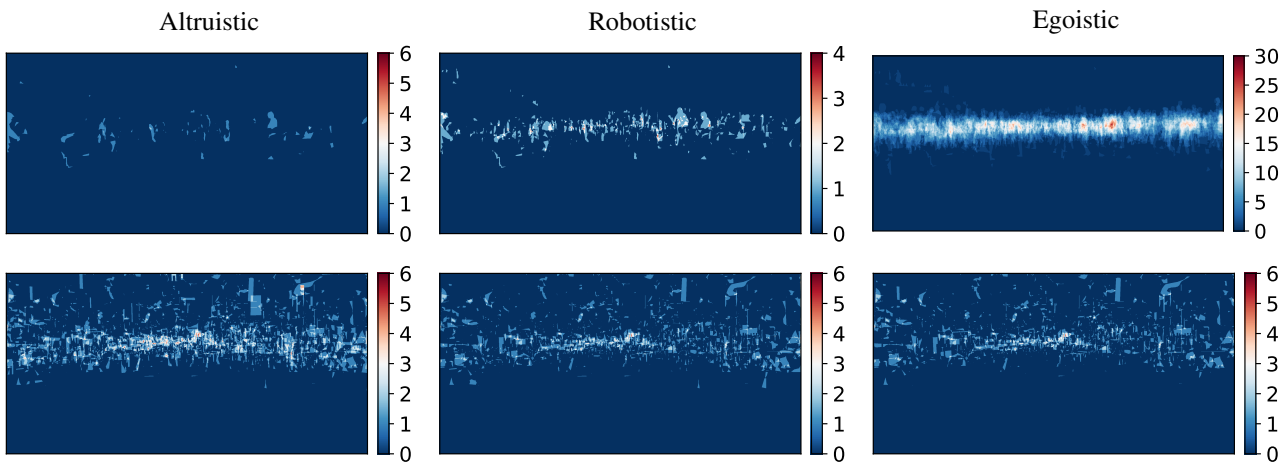

The behavior is different with respect to $C_A$. Regarding building segments, $C_R$ performs better in both metrics in the frontal RoI. The recall is reduced by $1.29$ and the precision by significant $58,43$ percent points. In the sidewalk RoI, the recall of $C_A$ is slightly improved ($0.64\%$) but the precision is again drastically reduced to $24.94\%$.
Noteworthy from this analysis is that DeepLabv3+ has difficulties in detecting separated ground truth segments of building instances which arise from objects in front of buildings and splitting the instance's actual connected component in the ground truth, see also \cref{fig:fn}.


\paragraph{Segment-wise false-detection vs. non-detection.}
\commentRC{Another interesting quantity are the entire false-detections and non-detections of person and building segments when using the different cost matrices.
In this regard, we now define \commentMR{a} segment to be, depending on the considered prediction or ground truth mask, a false positive / negative if the segment's intersection over union ($\textit{IoU}$) equals $0$. \Cref{fig:fp} and \cref{fig:fn} visualize the segments with $IoU=0$ in the prediction mask and ground truth mask, 
respectively, 
again for the classes person and building.}
The presented heatmaps visibly confirm the findings from the precision and recall analysis. The application of cost-based decision rules change\commentMR{s} the perception of DeepLabv3+ significantly. 
\commentRC{
For instance, for the class person the altruistic cost matrix overproduces
false positives
but \commentMR{there are almost no overlooked} person segments. On the contrary, the egoistic cost matrix almost completely refuses to predict the class person but is mostly correct in case it predicts a person segment.}
The robotistic cost matrix offers a balanced compromise between both prediction mistakes. Depending on people's individual sense of how the cost matrix should be defined, the presented observations will change again. Thus, what will remain open is a concrete suggestion to the inevitable definition of a confusion cost matrix.

\section{Discussion}

In this paper we illustrated the impact of cost-based decision rules on the perception of a state-of-the-art semantic segmentation neural network. In this framework, we discussed options for setting up cost-based decision rules ranging from the classical ``robotistic'' maximum a-posteriori probability principle over probability thresholding and the Maximum Likelihood decision rule to \emph{ad hoc} ``egoistic'' and ``altruistic'' cost assignments to confusion events. Within the triangle of robotistic, egoistic and altruistic attitudes, we investigated precision and recall and also false positive and negative rates in two regions of interest for the classes ``person'' and ``building'' in the Cityscapes dataset. We demonstrated the metrics' dependence on the convex combination of the cost matrices from the three mentioned ethical attitudes spanning a triangle within a larger space of values.

\commentHG{On the technical side, many questions concerning the use of cost-based decision rules have to be clarified, \emph{e.g.}~the adaptation of cost matrices to prior probabilities
or the impact on ``downstream'' modules like data fusion with other sensors and trajectory planning.}

\commentRD{Let us turn to the ethical side of the discussion. The probabilistic nature of the output of the segmentation network makes a decision rule necessary. As different decision rules have non-converging consequences, a choice for a decision rule amounts to a choice where in the long run human lives are weighted against other considerations. This choice is therefore not one to be made from a purely technical side (by \emph{e.g.}~choosing the mathematically ``natural'' decision rule) but one that needs to recognize its ethical dimension. While technological advances may have an impact on these considerations they will not make the need for a decision rule obsolete.}   

\commentRD{This leads to the question: Which decision rule is the ``right'' one? As in most cases of moral uncertainty, different normative ethical schools of thought will provide different answers (see \cite[Ch.3]{lin2014robot} for a short non-technical introduction in the context of robot ethics). A deontological strategy would try to justify a certain choice of a decision rule by arguing for the rule itself being ethically ``good'', not considering what may follow from that choice. For instance, a strict rule-based implementation of the requirement by the ethics commission that ``[t]he protection of individuals takes precedence over all other utilitarian considerations.''~\cite{ Ethikkommission} may be interpreted to lead to a cost function that is never allowed to confuse a human for another object. A consequentialist strategy justifies a cost function by focusing on the consequences of a certain choice. This would involve the above analysis of the consequences of the egoistic and altruistic cost functions. Another approach refers to polling, using the ethical intuition of the majority of the people being asked. This can lead to strong cultural differences, as resulted in an analysis of Awad et al. in the context of trolley-like problems \cite{Awad18}.}

\commentRD{It is not the aim of this paper to defend any specific approach or to provide an alternative answer to the above problem of choosing the ``right'' decision rule, but to make transparent the underlying ethical dimension of what may seem as mathematically innocuous ``natural'’ choices. This transparency is a precondition for a responsible handling and open debate on these issues.
}

\noindent \textbf{Acknowledgment.} R.~Chan, M.~Rottmann and H.~Gott-schalk acknowledge (partial) funding by Volkswagen Group Research through the contract ``Maximum likelihood and cost-based decision rules in semantic segmentation''.

{\small
\bibliographystyle{ieee}
\bibliography{lit}
}

\end{document}